\pdfoutput=1

\documentclass[11pt]{article}

\usepackage{acl}

\usepackage{times}
\usepackage{latexsym}

\usepackage[T1]{fontenc}
\usepackage{mathtools}
\usepackage{amsmath}
\usepackage{amssymb}
\usepackage{enumerate}
\usepackage{subfigure}
\usepackage{xspace}
\usepackage[font=small]{caption} 

\usepackage{booktabs}
\usepackage{graphicx}
\usepackage{multirow}

\usepackage[utf8]{inputenc}

\usepackage{microtype}

\newcommand*{\yoruba}{Yor\`ub\'a\xspace}

\title{Is BERT Robust to Label Noise? A Study on Learning with Noisy Labels in Text Classification}

\author{Dawei Zhu, Michael A. Hedderich, Fangzhou Zhai, David Ifeoluwa Adelani \& Dietrich Klakow \\
Saarland University, Saarland Informatics Campus, Germany \\
\texttt{\{dzhu,mhedderich,didelani,dietrich.klakow\}@lsv.uni-saarland.de} \\
\texttt{fzhai@coli.uni-saarland.de}}

\begin{document}
\maketitle
\begin{abstract}
Incorrect labels in training data occur when human annotators make mistakes or when the data is generated via weak or distant supervision. It has been shown that complex noise-handling techniques - by modeling, cleaning or filtering the noisy instances - are required to prevent models from fitting this label noise. 
However, we show in this work that, for text classification tasks with modern NLP models like BERT, over a variety of noise types, existing noise-handling methods do not always improve its performance, and may even deteriorate it, suggesting the need for further investigation. We also back our observations with a comprehensive analysis.
\end{abstract}

\section{Introduction}
For many languages, domains and tasks, large datasets with high-quality labels are not available. To tackle this issue, cheaper data acquisition methods have been suggested, such as crowdsourcing or automatic annotation methods like weak and distant supervision. Unfortunately, compared to gold-standard data, these approaches come with more labeling mistakes, which are known as noisy labels. Noise-handling has become an established approach to mitigate the negative impact of learning with noisy labels. A variety of methods have been proposed that model the noise, or clean and filter the noisy instances \cite{hedderich-etal-2021-survey, DBLP:journals/kbs/AlganU21}. \citet{DBLP:conf/naacl/JindalPLN19} show e.g. a 30\% boost in performance after applying noise-handling techniques on a CNN-based text classifier.

In a recent work, \citet{DBLP:journals/corr/abs-2105-00828} showed that BERT \cite{DBLP:conf/naacl/DevlinCLT19} has an inherent robustness against noisy labels. The generalization performance on the clean distribution drops only slowly with the increase of the mislabeled samples. Also, they show that early-stopping is crucial for learning with noisy labels as BERT will eventually memorize all wrong labels when trained long enough. However, their experiments only focus on a single type of noise and a limited range of noise levels. It remains unclear if BERT still performs robustly under a wider range of noise types and with higher fractions of mislabeled samples. Moreover, they perform early-stopping on a clean validation set, which may not be available under low resource settings. Last but not least, they do not compare to any noise-handling methods. 

\begin{figure}[]
    \centering
    \includegraphics[width=0.4\textwidth]{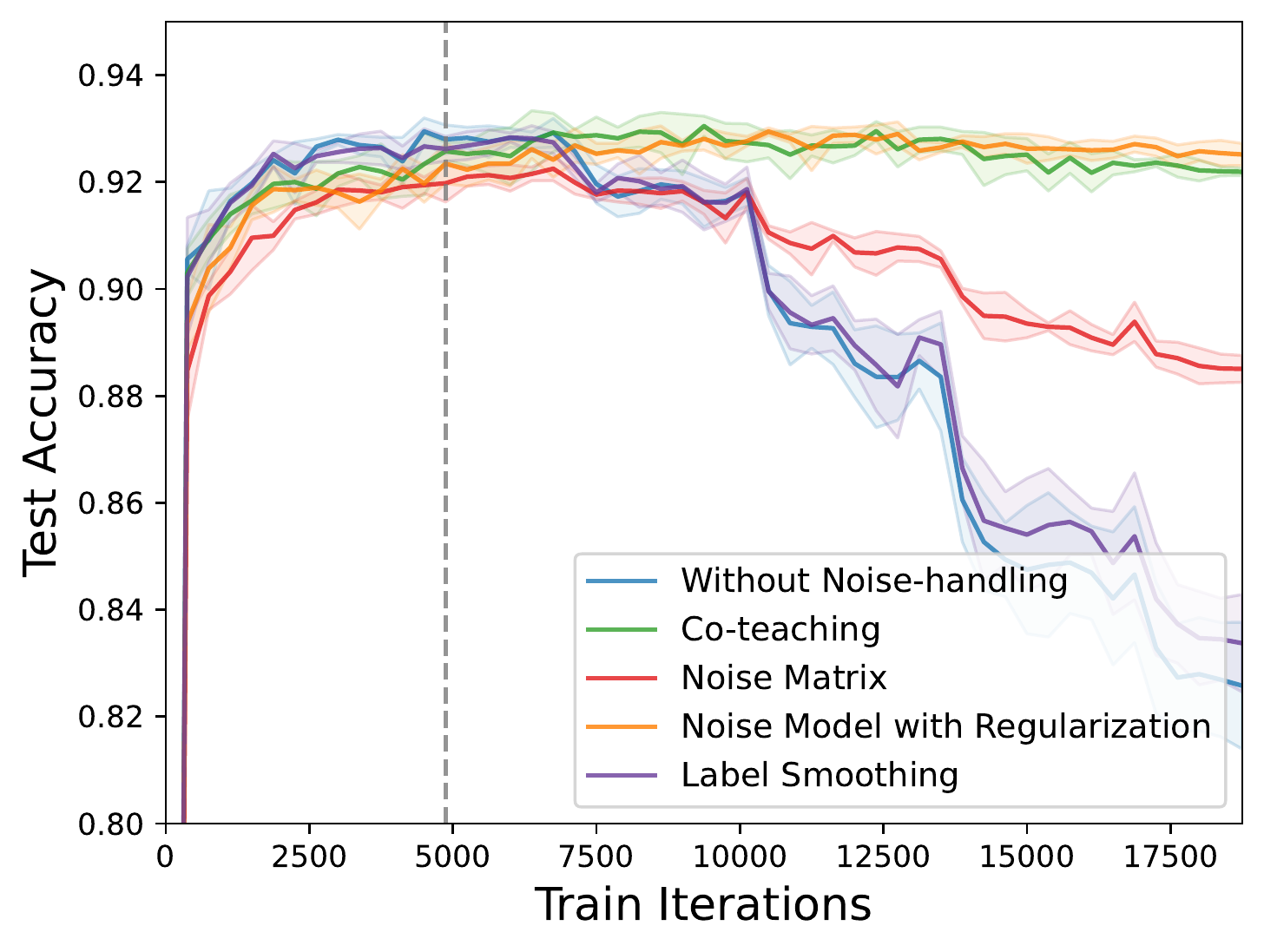}
    \caption{A typical training curve when learning with noise. Learning without noise-handling (blue) will reach a peak accuracy before memorizing the noise. Early-stopping on a noisy validation set (vertical grey line) is often sufficient to find such a peak. Injected uniform noise of 40\% on AG-News dataset. }
    \label{fig:introduction_fig}
\end{figure}

In this work, we investigate the behaviors of BERT on tasks with different noise types and noise levels. We also study the effect of noise-handling methods under these settings. Our main results include \textbf{(1)} BERT is robust against injected noise, but could be vulnerable to noise from weak supervision. In fact, the latter, even at a low level, can be more challenging than high injected noise. \textbf{(2)} Existing noise-handling methods do not improve the peak performance of BERT under any noise settings we investigated; as is shown with further analysis, noise-handling methods rarely render the correct labels more distinguishable from the incorrect ones.
\footnote{Our implementation is available on: \url{https://github.com/uds-lsv/BERT-LNL}.}

\section{Learning with Noisy Labels}
\paragraph{Problem Settings} We consider a $k$-class classification problem. Let $D$ denote the true data generation distribution over $\mathcal{X}\times\mathcal{Y}$ where $\mathcal{X}$ is the feature space and $\mathcal{Y}=\{1,...,k\}$ is the label space. In a typical classification task, we are provided with a training dataset $S=\{(x_i, y_i)_{i=1}^n\}$ sampled from $D$. In learning with noisy labels, however, we have no access to $D$. Instead, a noisy training set $\hat{S}=\{(x_i, \hat{y}_i)_{i=1}^n\}$ sampled from a label-corrupted data distribution $\hat{D}$. The goal is to learn a classifier that generalizes well on the clean distribution by only exploiting $\hat{S}$.
\paragraph{Injected Label Noise} To rigorously evaluate noise-handling methods at different noise levels, researchers in this area often construct noisy datasets from clean ones by injecting noise. This can, e.g., reflect annotation scenarios such as crowdsourcing, where some annotators answer randomly or prefer an early entry in a list of options. Modeling such noise is achieved by flipping the labels of the clean instances according to a pre-defined noise level $\varepsilon\in[0,1)$ and a noise type. There are two commonly used noise types: the single-flip noise \cite{DBLP:journals/corr/ReedLASER14}:

\begin{small}
\begin{equation*}
    p_{\text{sflip}}(\hat{y}=j|y=i) = \left\{\begin{array}{lr}
        1-\varepsilon, & \text{for } i = j \\
        \varepsilon, & \text{for one } i \neq j\\
        0, & else
        \end{array} \right. \;
\end{equation*}
\end{small}
and uniform-flip \cite{DBLP:conf/nips/RooyenMW15} noise

\begin{small}
\begin{equation*}
    p_{\text{uni}}(\hat{y}=j|y=i) = \left\{\begin{array}{lr}
        1-\varepsilon, & \text{for } i = j\\
        \frac{\varepsilon}{k-1}, & \text{for } i \neq j
        \end{array} \right. \; .
\end{equation*}
\end{small}

These noise generation processes are feature-independent, i.e. $p(\cdot|y=i,x) = p(\cdot|y=i)$. Therefore, they can be described by a noise transition matrix $T$ with $T_{ij} \coloneqq p(\hat{y}=j|y=i)$. It is usually assumed that the noise is diagonally-dominant when generating the noisy labels, i.e. $\forall i, T_{ii}>max_{j\neq i}T_{ij}$. 

\paragraph{Label Noise from Weak Supervision} Distant and weak supervision \cite{mintz-etal-2009-distant, NIPS2016_6709e8d6} have become essential methods to acquire labeled data in low-resource scenarios. The resulting noise, unlike injected noise, is often feature-dependent \cite{lange-etal-2019-feature}. We evaluate our methods on two real-world datasets in Hausa and \yoruba to cover this type of noise.

\section{Early-Stopping on Noisy Validation Set}
When trainied on noisy data without noise-handling, BERT reaches a high generalization performance before it starts fitting the noise. Then it memorizes the noise and the performance on clean distribution drops dramatically (the blue curve in Figure \ref{fig:introduction_fig}). Hence, for models without noise-handling, it is crucial to stop training when the generalization performance reaches its maximum. 

\citet{DBLP:journals/corr/abs-2105-00828} use a clean validation set to find this point. However, a clean validation set is often unavailable in realistic low-resource scenarios as it requires manual annotation. Therefore, we use a noisy validation set for early-stopping in all of our experiments and we attain models that generalize well on the clean distribution.

In our example in Figure \ref{fig:introduction_fig}, we see that while most noise-handling methods prevent BERT from fitting the noise in the long run, their peak performance is not significantly higher than a vanilla model without noise-handling.

\begin{table}[]
\centering\scriptsize
\begin{tabular}{@{}c@{\hspace{0.75\tabcolsep}}c@{\hspace{0.75\tabcolsep}}c@{\hspace{0.75\tabcolsep}}c@{\hspace{0.75\tabcolsep}}c@{\hspace{0.75\tabcolsep}}c@{\hspace{0.75\tabcolsep}}c@{}}
\toprule
Dataset & Classes & \begin{tabular}[c]{@{}c@{}}Average\\ Lengths\end{tabular} & \begin{tabular}[c]{@{}c@{}}Train\\ Samples\end{tabular} & \begin{tabular}[c]{@{}c@{}}Validation\\ Samples\end{tabular} & \begin{tabular}[c]{@{}c@{}}Test \\ Samples\end{tabular} & \begin{tabular}[c]{@{}c@{}}Train\\ Noise Level\end{tabular} \\ \midrule
IMDB & 2 & 292 & 21246 & 3754 & 25000 & various \\
AG-News & 4 & 44 & 108000 & 12000 & 7600 & various \\
\yoruba & 7 & 13 & 1340 & 189 & 379 & 33.28\% \\
Hausa & 5 & 10 & 2045 & 290 & 582 & 50.37\% \\ \bottomrule
\end{tabular}
\caption{Statistics of the text classification datasets. The train noise level is the false discovery rate  (i.e. 1-precision) of the noisy labels in the training set. %
The original AG-News has 120k training instances and no validation instances. We therefore held-out 10\% of the training samples for validation. %
}
\label{tab:datasets}
\end{table}

\section{Experiments}
\label{section: experiments}
\paragraph{Dataset Construction}
We experiment with four text classification datasets: two benchmarks, AG-News \cite{DBLP:conf/nips/ZhangZL15} and IMDB \cite{DBLP:conf/acl/MaasDPHNP11}, injected with different levels of single-flip or uniform noise; for the weakly supervised noise, we make use of two news topics datasets in two low-resource languages: Hausa and \yoruba ~\cite{DBLP:conf/emnlp/HedderichAZAMK20}. Hausa and \yoruba
are the second and the third most spoken indigenous language in Africa, with 40 and 35 million native speakers, respectively \cite{Ethnologue2019}. The noisy labels were gazetteered. For example, to identify texts for the class ``Africa'', a labeling rule based on a list of African countries and their capitals is used. Note that while we can vary the noise levels of injected noise, the amount of weak supervision noise in Hausa and \yoruba is fixed\footnote{refer to Appendix \ref{sec:appendix_nm_yh} for detailed noise distribution.}. We summarize some basic statistics of the datasets in Table \ref{tab:datasets}.

\paragraph{Implementation} We use of-the-shelf BERT models for our tasks. Specifically, we apply the BERT-base model for AG-News and IMDB, and the mBERT-base for \yoruba and Hausa. The fine-tuning approach follows \cite{DBLP:conf/naacl/DevlinCLT19}. In all settings, we apply early-stopping on a noisy validation set to mimic the realistic low-resource settings, while the test set remains clean. For more implementation details and a discussion on clean and noisy validation sets, see Appendix \ref{sec: appendix_compare_ec_en} and \ref{sec: appendix_implementation_detail}.

\subsection{Baselines}

We compare learning without noise-handling with four popular noise-handling methods.\footnote{For a fair comparison, early-stopping on a noisy validation set is applied to all four noise-handling methods.}

\paragraph{Without Noise-handling} Train BERT on the noisy training set as it was clean. A noisy validation set is used for early-stopping. 

\paragraph{No Validation} For the sake of comparison, we train the model without noise-handling and until the training loss converges. 

\paragraph{Noise Matrix} A noise transition matrix is appended after BERT's prediction to transform the clean label distribution to the noisy one. A variety of methods exists for estimating the noise matrix, i.e. \citet{sukhbaatar2015training, DBLP:conf/icassp/BekkerG16, DBLP:conf/cvpr/PatriniRMNQ17, DBLP:conf/nips/HendrycksMWG18, DBLP:conf/nips/YaoL0GD0S20}. To exclude the effects of estimation errors in the evaluation, we use the ground truth transition matrix as it is the best possible estimation. This matrix is fixed after initialization.

\paragraph{Noise Matrix with Regularization} The previous state-of-the-art for text classification with noisy labels \cite{DBLP:conf/naacl/JindalPLN19}. Similar to \textit{Noise Matrix}, it appends a noise matrix after BERT's output. During training, the matrix is learned with an $l2$ regularization and is not necessarily normalized to be a probability matrix. In the original implementation they use CNN-based models as backbone, we switch it to BERT for fair comparison.

\paragraph{Co-teaching} \citet{DBLP:conf/nips/HanYYNXHTS18} Train two networks to pick cleaner training subsets for each other. The Co-teaching framework requires an estimation of the noise level. Similarly to NMat, we use the ground truth noise level to exclude the performance drop caused by estimation error. 

\paragraph{Label Smoothing} Label smoothing \cite{Szegedy2016RethinkingTI} is a commonly used method to improve model's generalization and calibration. It mixes the one-hot label with a uniform vector, preventing the model from getting overconfident on the samples. \citet{lukasik2020does} further shows that it improves noise robustness.

\begin{figure*}[t]
    \centering
    \subfigure[AG-News, uniform noise]{
        \includegraphics[width=0.65\columnwidth]{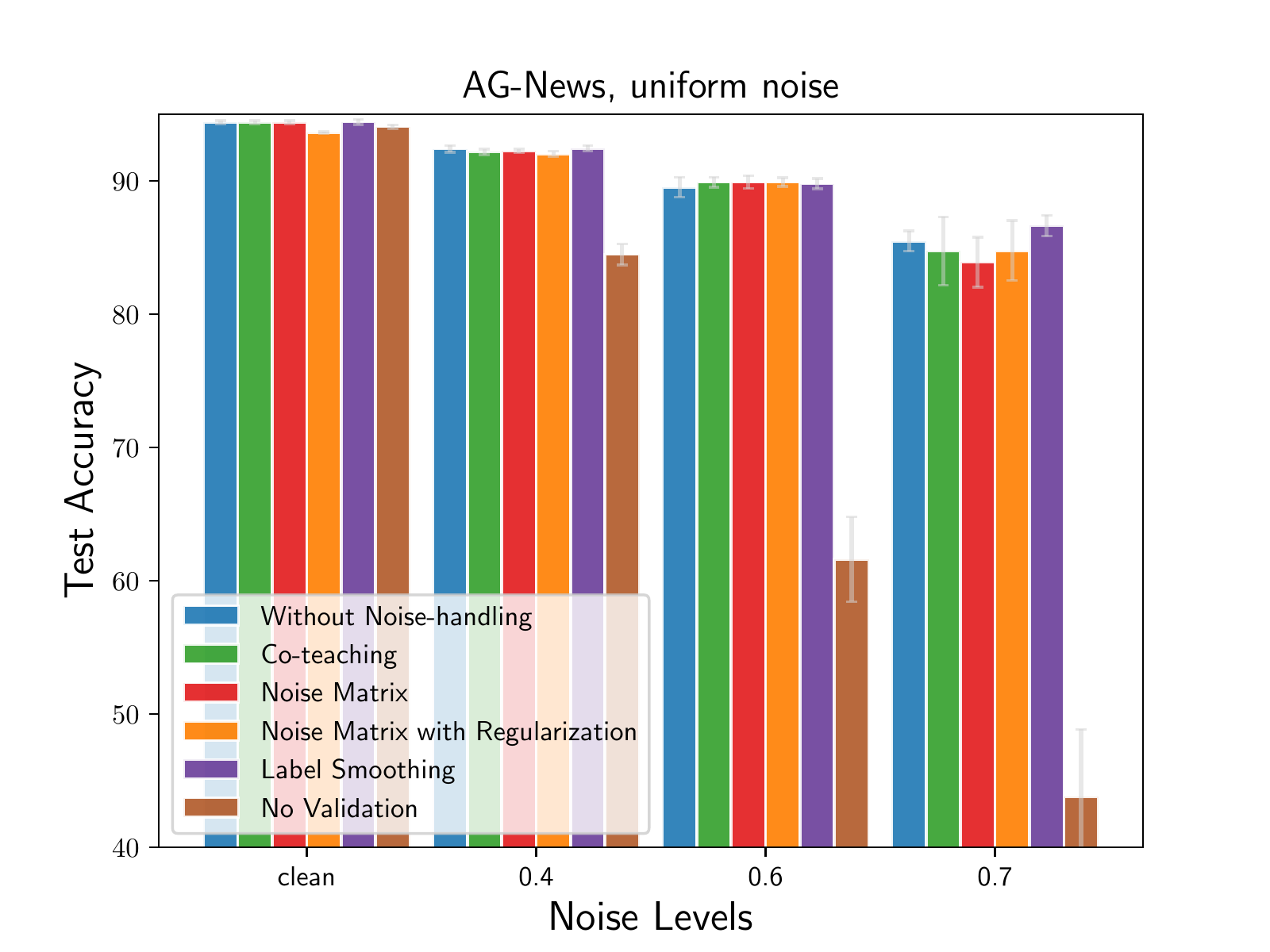}
        \label{subfig:agnews_uniform}
    }
    \subfigure[IMDB, single-flip noise]{
        \includegraphics[width=0.65\columnwidth]{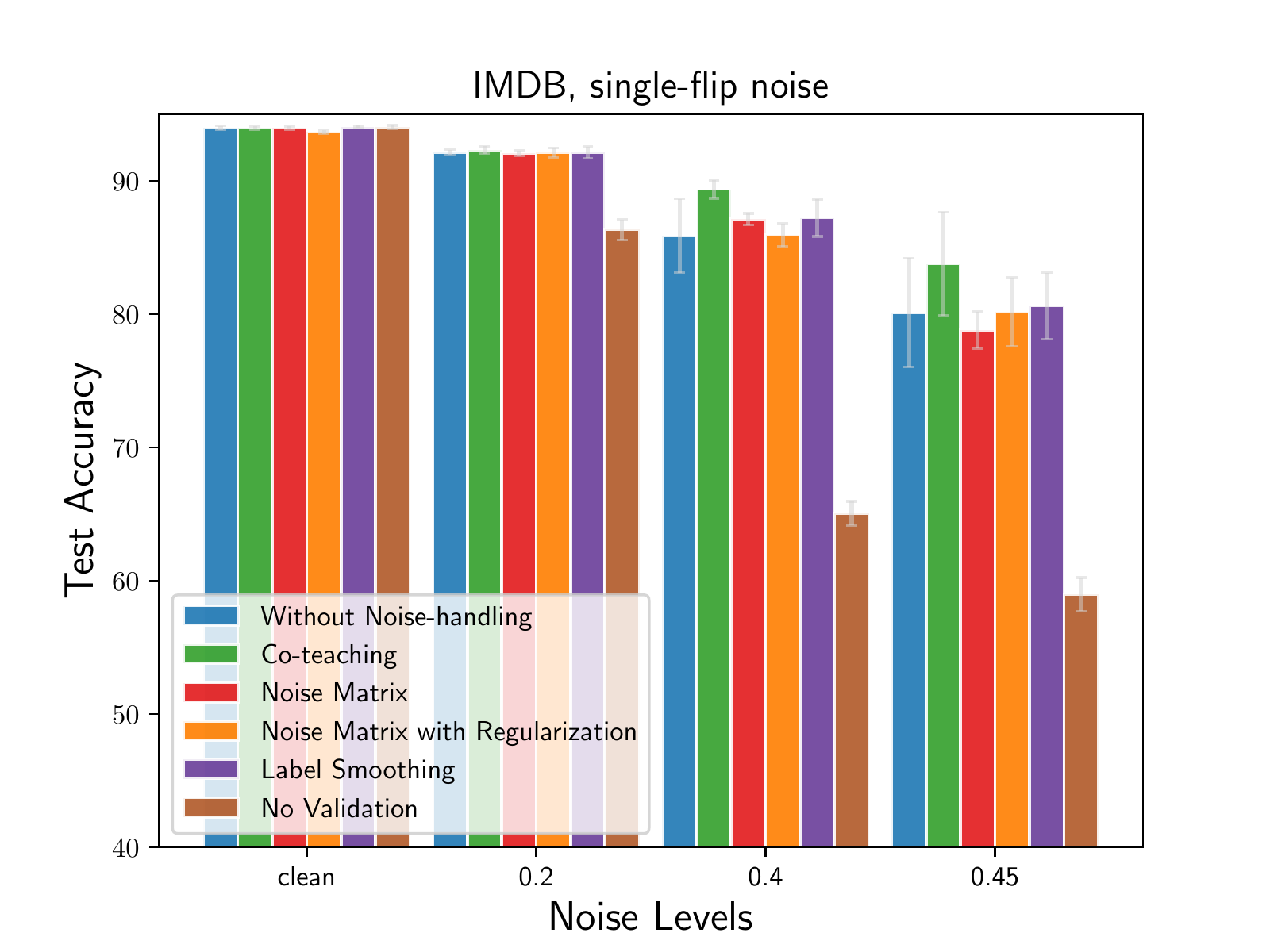}
    }
    \subfigure[\yoruba \& Hausa, weak supervision noise]{
        \includegraphics[width=0.65\columnwidth]{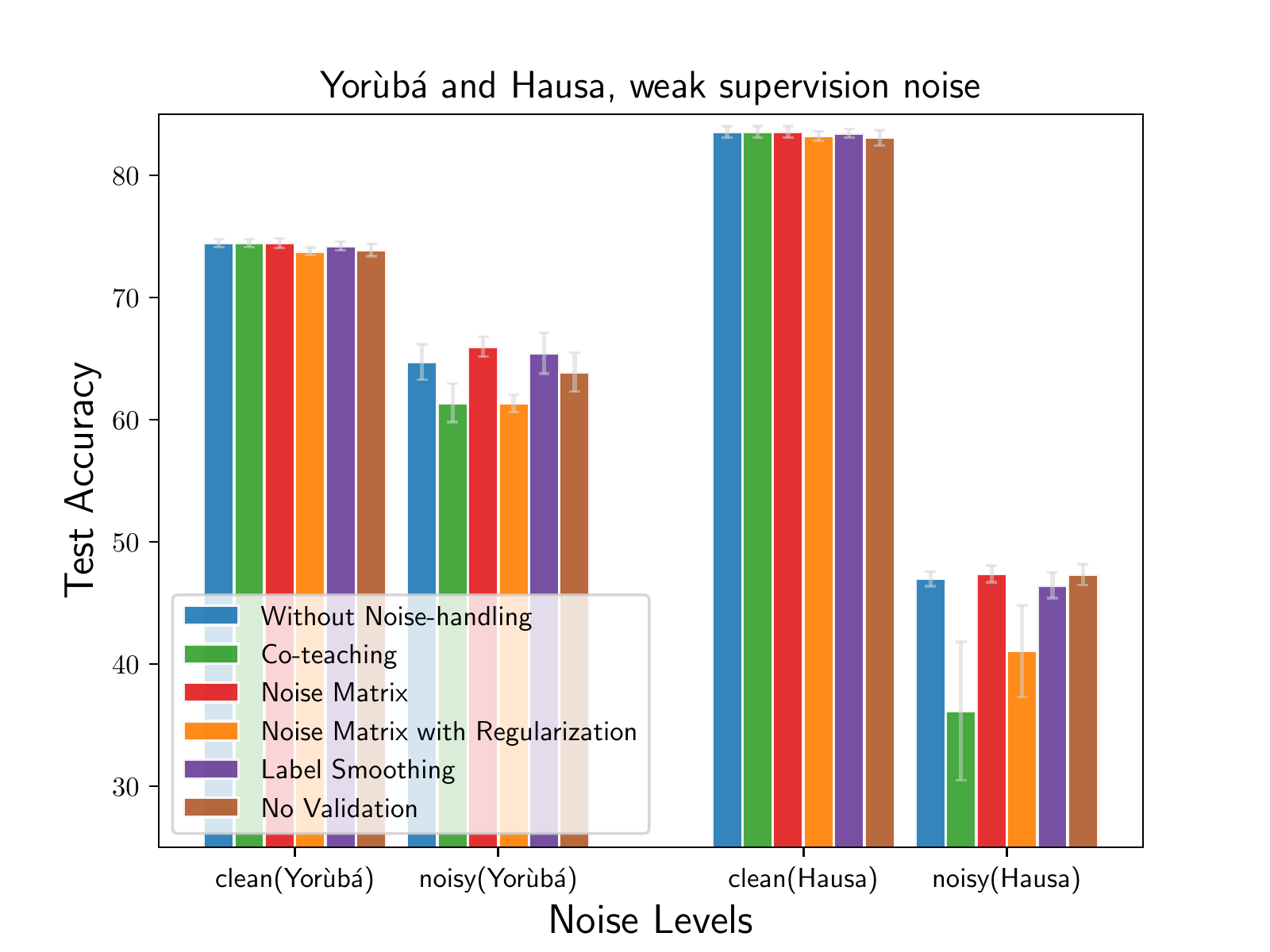}
        \label{subfig:yoruba_hausa_performance}
    }
    \vspace*{-4mm} 
    \caption{Test accuracy in different noise settings. a) \& b) injected noise with different noise levels c) weak supervision noise, at noise levels of 33.28\% and 50.37\% in \yoruba and Hausa, respectively. Noise-handling methods do not always improve peak performances. Further plots in Appendix \ref{sec:appendix_performance_table}.
    }
    \label{fig:performance_bar_plots}
\end{figure*}
\begin{figure*}[t]
    \centering
    \subfigure[AG-News - 70\% uniform noise]{
        \includegraphics[width=0.65\columnwidth]{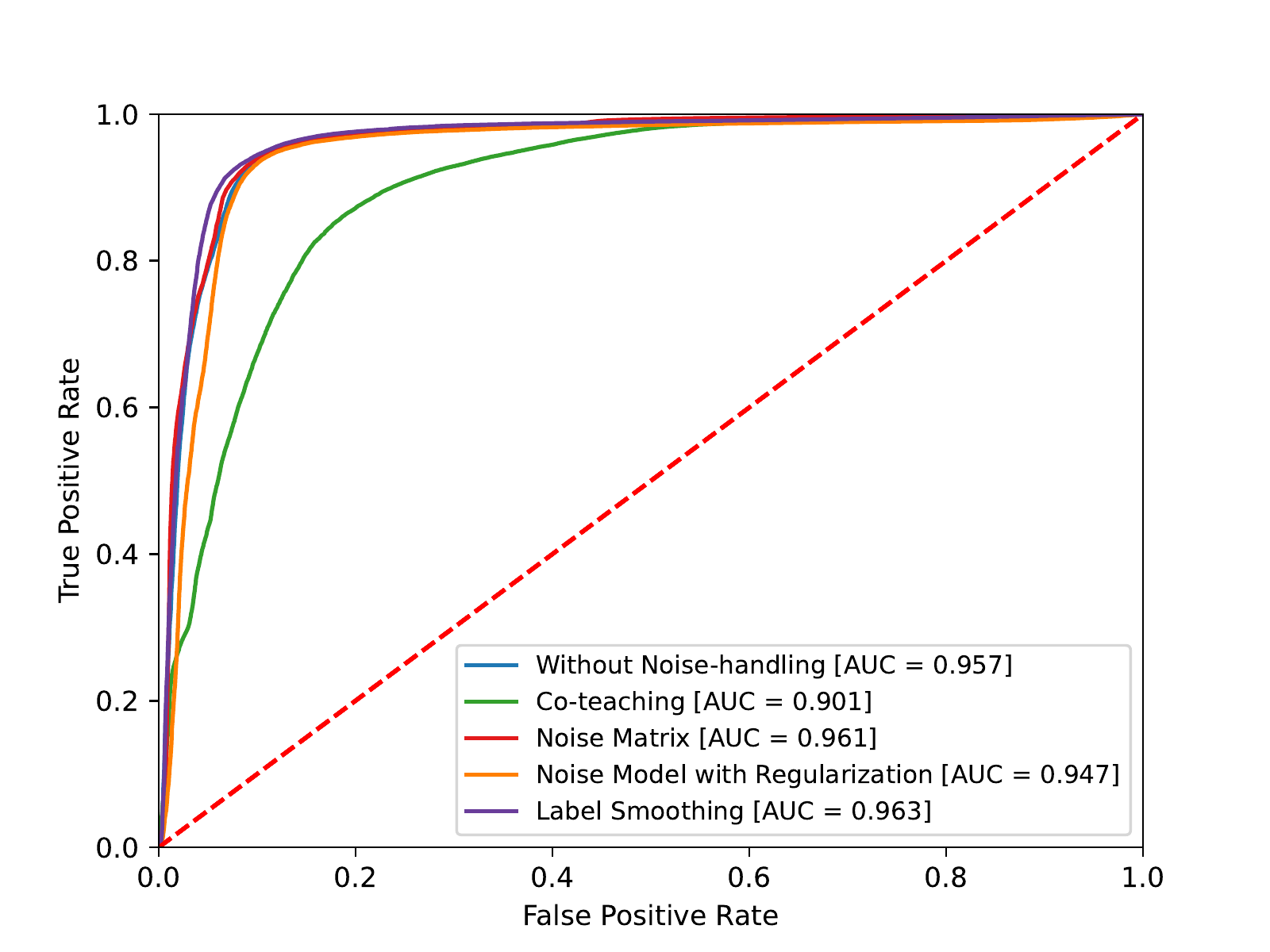}
        \label{fig:roc_plots_agnews_u70}
    }
    \subfigure[\yoruba$\;$- weak supervision noise]{
        \includegraphics[width=0.65\columnwidth]{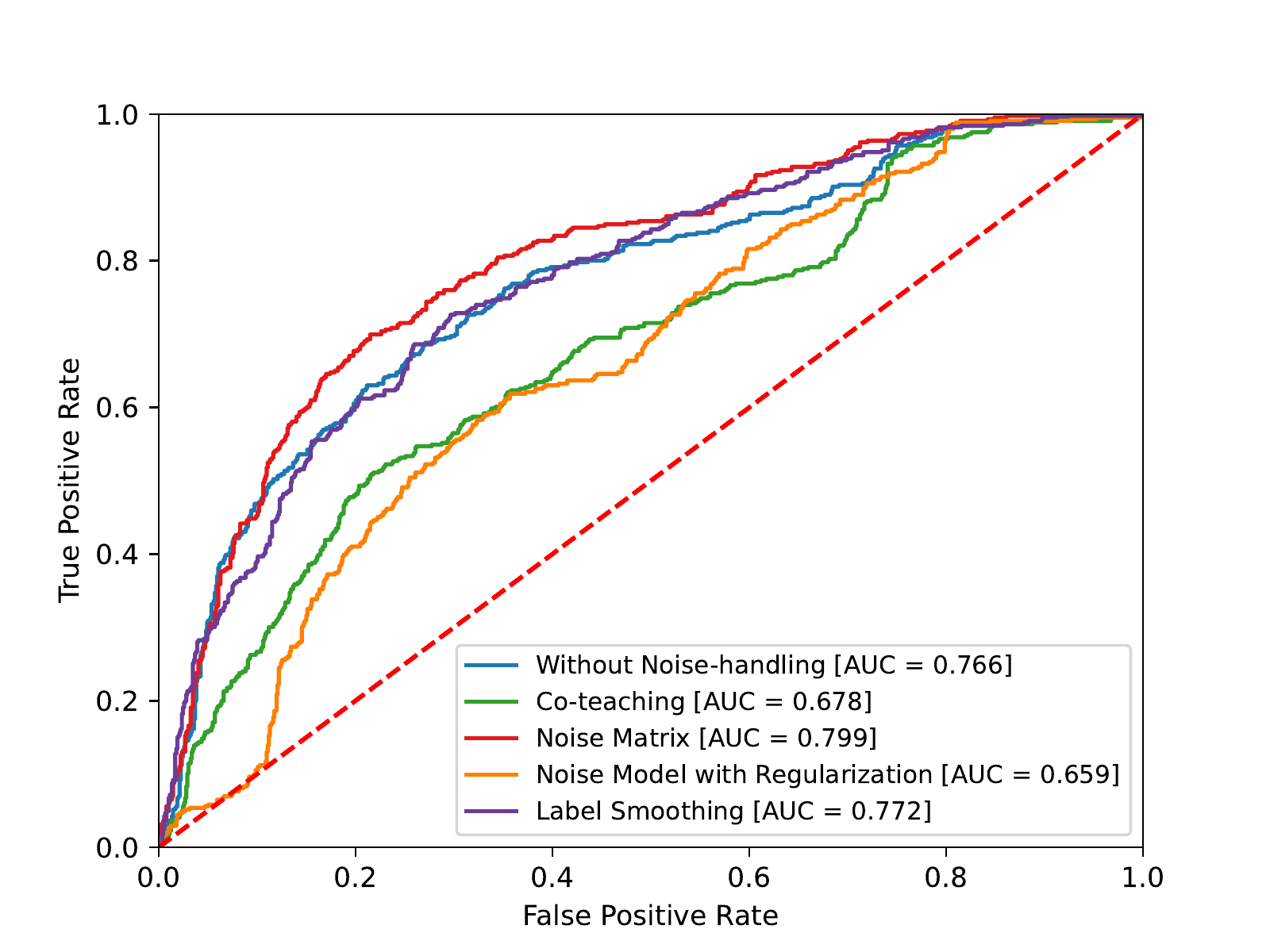}
    }
    \subfigure[Hausa - weak supervision noise]{
    \includegraphics[width=0.65\columnwidth]{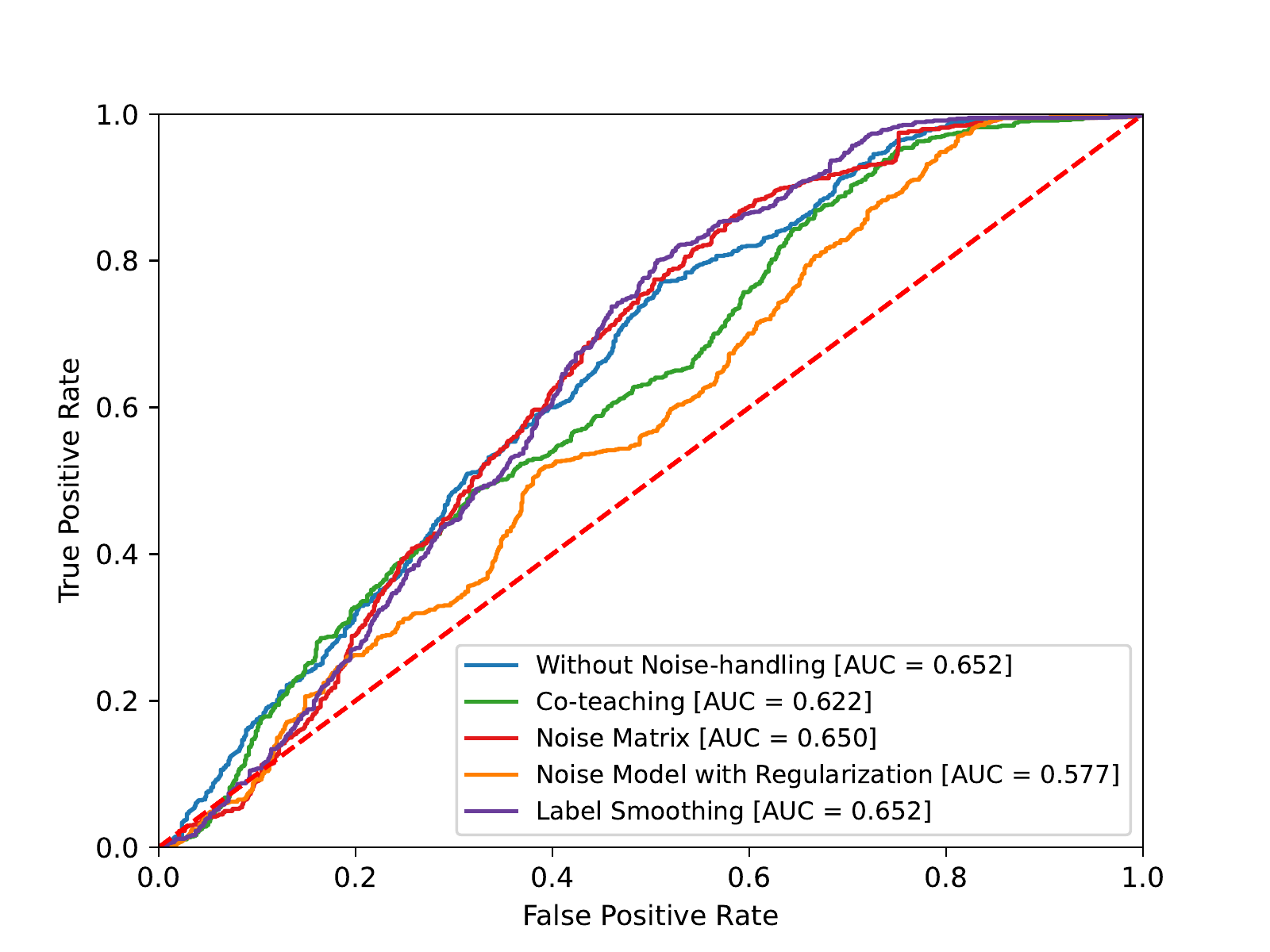}
    }
    \vspace*{-3mm} 
    \caption{ROC curves on wrong label detection (binary classification) using the losses. The losses are recorded at the training step when early-stopping is triggered. Noise-handling methods do not make the losses of correct and incorrect labels more distinguishable.
    Further plots in Appendix \ref{sec:appendix_more_roc_curves}. }
    \label{fig:roc_plots}
\end{figure*}

\subsection{Experimental Results}
\label{sec:main_results}
We evaluate our baselines on both injected noise (on AG-News and IMDB) and weak supervision noise (on Hausa and \yoruba). The test accuracy is presented Figure \ref{fig:performance_bar_plots}. On injected noise, our results match and extend the findings by \citet{DBLP:journals/corr/abs-2105-00828} that BERT is noise robust. For example, the test accuracy drops only about 10\% after injecting 70\% wrong labels (Figure \ref{subfig:agnews_uniform}). However, we find that BERT is vulnerable under weak supervision noise. The performance can drop up to 35\% in a dataset like Hausa with 50\% weak supervision noise compared to training with clean labels (Figure \ref{subfig:yoruba_hausa_performance}). This indicates that the experience on injected noise may not be transferable to weak supervision noise. 

We also observe that noise-handling methods are not always helpful. For injected noise, the benefits from noise-handling become obvious only under high noise levels. But even then, there is no clear winner, meaning that it is hard to decide beforehand which noise method to apply - with the risk that they may even perform worse than BERT without noise-handling. The same applies to weak supervision noise. The maximal performance gap between the best model and BERT without noise-handling is less than 4\% and 1.5\% under injected noise and weak supervision noise, respectively.

\begin{figure}[t]
    \centering
    \includegraphics[width=0.35\textwidth]{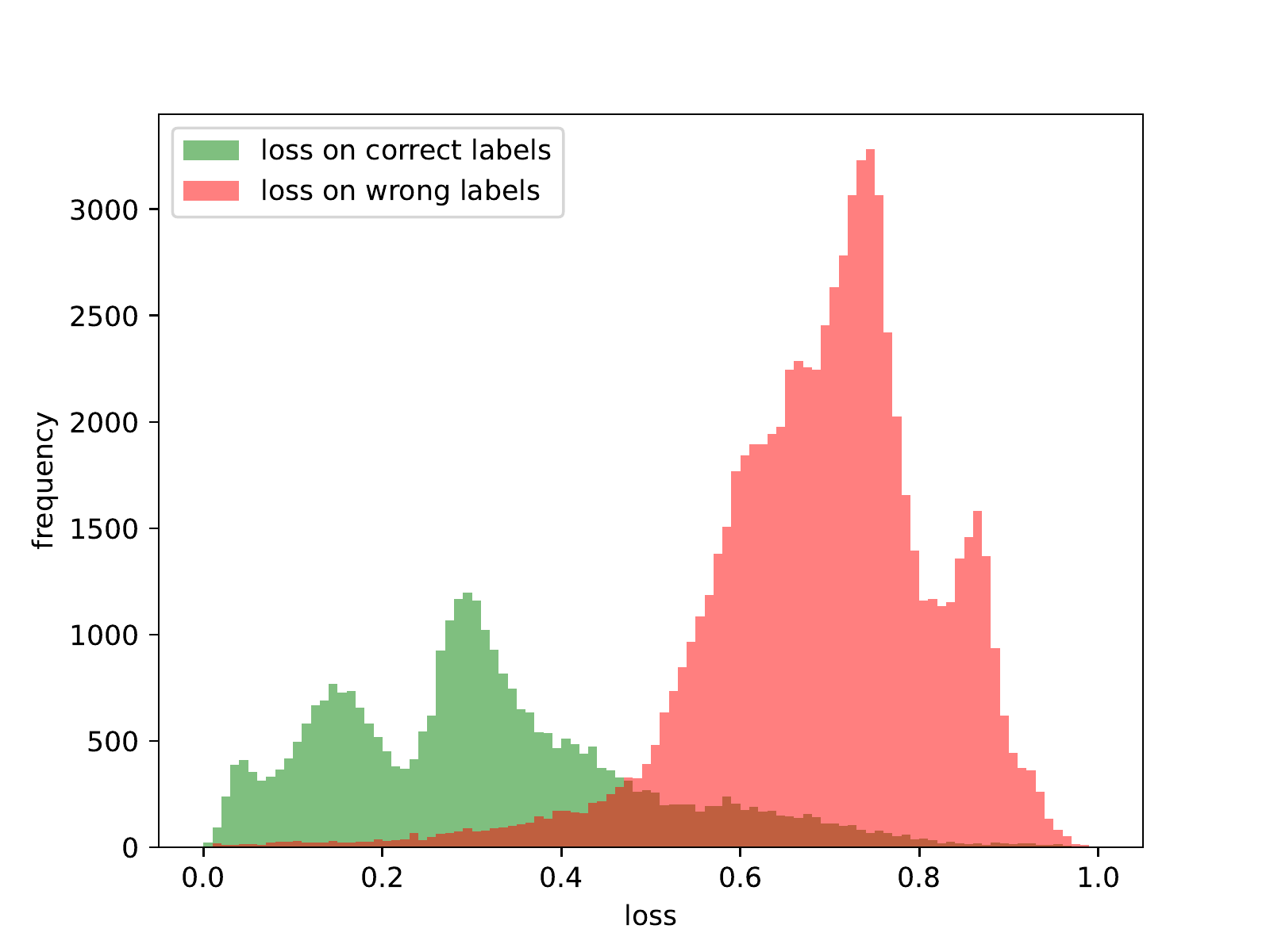}
    \vspace*{-1mm}
    \caption{Loss histogram at the training iteration when the early-stopping is triggered. AG-News dataset with 70\% uniform noise.}
    \label{fig:bert_learning_curve_example}
\end{figure}

\subsection{Analysis of Loss Distributions}
To shed some light on why BERT is robust against injected noise but not weak supervision noise, we track the losses on correctly and wrongly labeled samples during training. Figure \ref{fig:bert_learning_curve_example} depicts typical distributions of losses associated with correctly and incorrectly labeled samples, respectively, when early-stopping is triggered. We see that they have minimal overlap, thus different behaviors throughout the training, potentially allowing the model to distinguish correctly and incorrectly labeled samples from each other. We could further quantify the difference by their separability. Figure \ref{fig:roc_plots} presents the receiver operating characteristic (ROC) curves of a thresholds-based classifier. We observe that \textbf{(1)} under injected noise, an area under curve (AUC) of more than 90 can be easily achieved without noise-handling (Figure \ref{fig:roc_plots_agnews_u70}), supporting our observation that injected noise has rather a low impact on BERT. \textbf{(2)} Under weak-supervision noise, the AUC score is significantly lower, which means the correct and incorrect labels are less distinguishable. Therefore, BERT fits both labels at similar rates. One reason could be that the noise in weak supervision is often feature-dependent, it might become easier for BERT to fit them, which in turn deteriorates the generalization. \textbf{(3)} We do not observe a raise in AUC scores when applying noise-handling methods, indicating that noise-handling methods rarely enhance BERT's ability to further avoid the negative impact of wrong labels. This is consistent with the observation in Section \ref{sec:main_results} that noise-handling methods have little impact on BERT's generalization performance.

\section{Conclusion}
On several text classification datasets and for different noise types, we showed that BERT is noise resistant under injected noise, but not necessarily under weak supervision noise. In both cases, the improvement obtained by applying noise-handling methods are limited. Our analysis on the separability of losses corresponding to correct and incorrect labeled samples provides evidence to this argument. Our analysis offers both motivation and insights to further improve label noise-handling methods and make them useful on more realistic types of noise.

\section{Broader Impact Statement and Ethics}
Noisy labels are a cheaper source of supervision. This could make it easier to use machine learning for improper use cases. However, it also opens up NLP methods for low-resource settings such as under-resourced languages or applications developed by individuals or small organizations. It can, therefore, be a step towards the democratization of AI.

\section*{Acknowledgments}
This work has been partially funded by the Deutsche Forschungsgemeinschaft (DFG, German Research Foundation) – Project-ID 232722074 – SFB 1102 and the EU Horizon 2020 projects ROXANNE under grant number 833635 and COMPRISE under grant agreement No. 3081705.

\bibliography{acl2022_ref}
\bibliographystyle{acl_natbib}

\newpage\phantom{1}\newpage

\appendix

\section{Noise Matrix on \yoruba and Hausa}
\label{sec:appendix_nm_yh}
\begin{figure}[h]
    \centering
    \subfigure[\yoruba]{{\includegraphics[width=3.5cm]{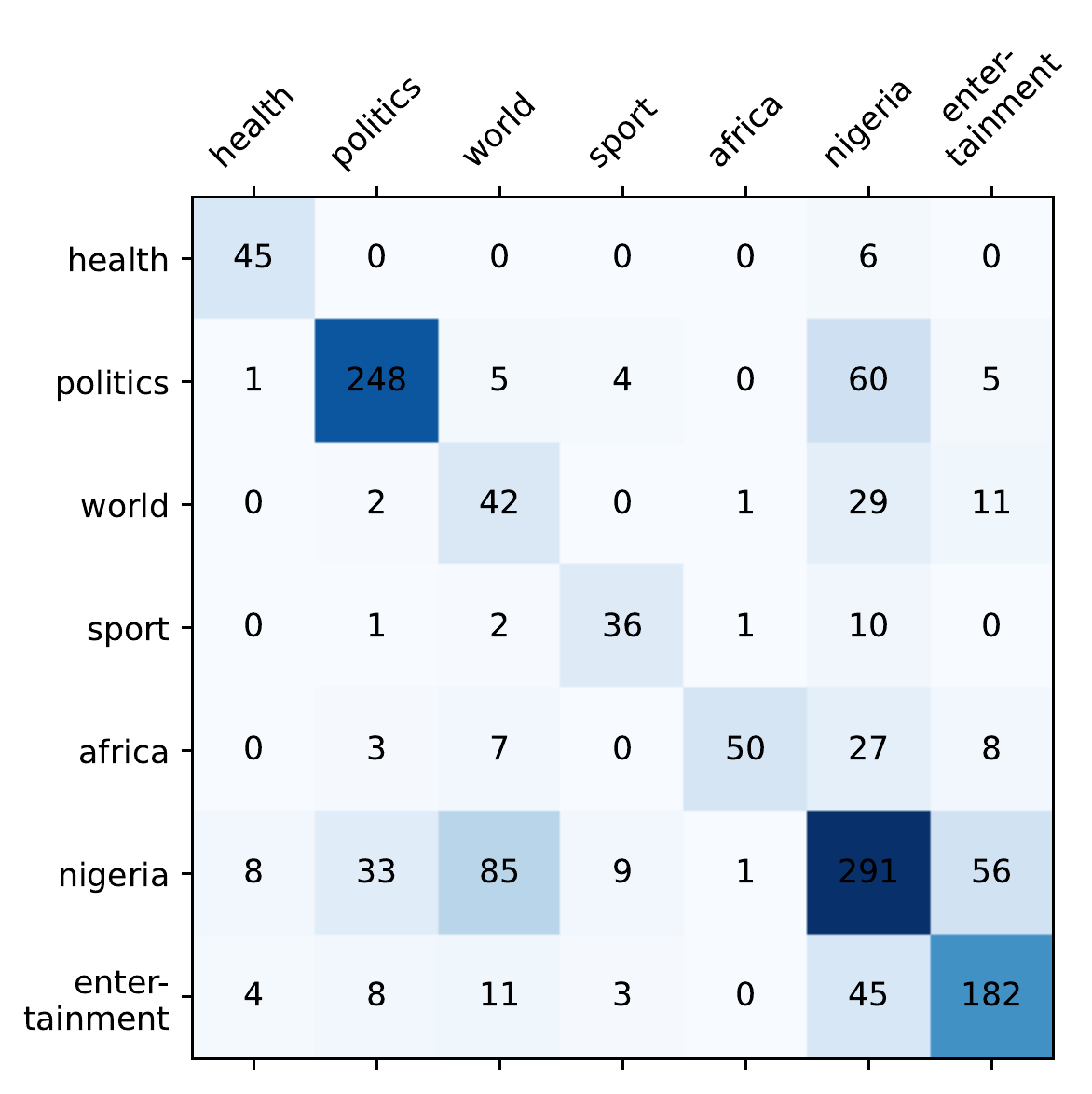} }}%
    \subfigure[Hausa]{{\includegraphics[width=3.5cm]{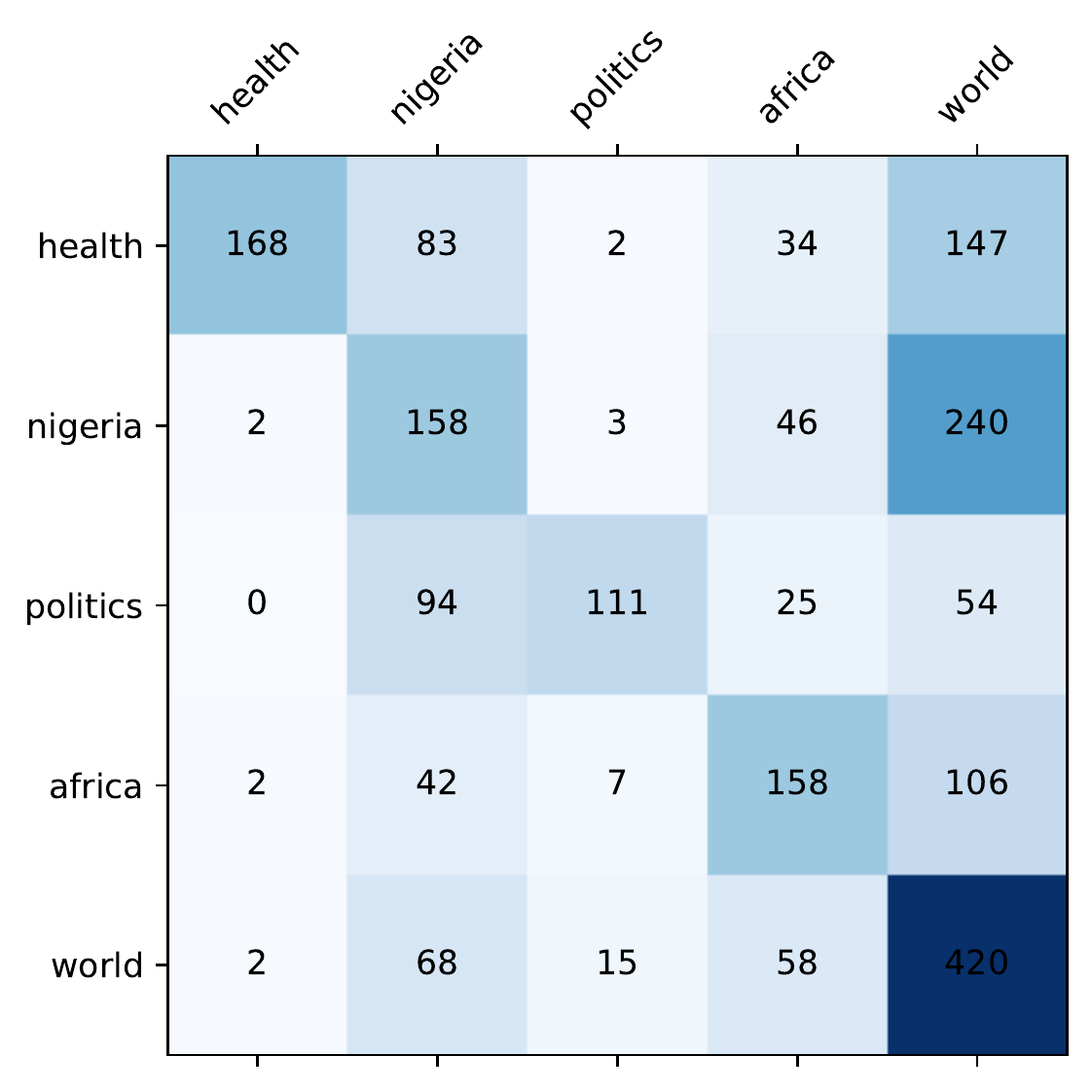} }}%
    \caption{Noise matrix constructed from the \yoruba (Hausa) training set.}%
    \label{fig:yoruba_hausa_mat}%
\end{figure}

The training and validation sets of \yoruba and Hausa have two sets of labels: the human-annotated (clean) labels and labels obtained from weak supervision. This makes it possible to compute the ground truth noise matrix in the training set. The noise matrices in \yoruba and Hausa are shown in Figure \ref{fig:yoruba_hausa_mat}. The Noise Matrix method evaluated in section \ref{section: experiments} uses these two matrices for initialization. The labeling rules in the weak supervision are described in \cite{DBLP:conf/emnlp/HedderichAZAMK20}. The \yoruba dataset has a rather low noise level, and the diagonally-dominant noise assumption holds in the training set. Oppositely, the Hausa training set is quite noisy. For the label ``nigeria'' the wrong labels is overwhelming, violating the diagonally-dominant noise assumption. Label ``politics'' is often misrecognized as ``nigeria''.  Moreover, many labels are misrecognized as the label ``world'', making an unbalanced classification dataset. These factors make it very challenging to conquer the noise in this dataset.

\begin{table*}[]
\centering\scriptsize
\scalebox{0.95}{
\begin{tabular}{@{}llllllllll@{}}
\toprule
\multicolumn{1}{c}{} & \multicolumn{6}{c}{AG-News} & \multicolumn{3}{c}{IMDB} \\ \cmidrule(lr){2-7} \cmidrule(lr){8-10}
 & \multicolumn{3}{c}{uniform} & \multicolumn{3}{c}{single-flip} & \multicolumn{3}{c}{single-flip} \\ \cmidrule(lr){2-4} \cmidrule(lr){5-7} \cmidrule(lr){8-10}
  & \multicolumn{1}{c}{40\%} & \multicolumn{1}{c}{60\%} & \multicolumn{1}{c}{70\%} & \multicolumn{1}{c}{20\%} & \multicolumn{1}{c}{40\%} & \multicolumn{1}{c}{45\%} & \multicolumn{1}{c}{20\%} & \multicolumn{1}{c}{40\%} & \multicolumn{1}{c}{45\%}\\ \midrule
Performance Difference (\%) & 0.10$\pm$0.09 & 0.56$\pm$0.50 & 1.96$\pm$0.97 & 0.06$\pm$0.07 & 0.29$\pm$0.19 & 2.00$\pm$0.60 & 0.14$\pm$0.19 & 1.71$\pm$2.05 & 1.76$\pm$2.79 \\ \bottomrule
\end{tabular}}
\caption{Average performance difference (in \%) and standard deviation (5 trials) between the test accuracy based on early-stopping with the clean validation and with the noisy validation set.}
\label{tab:simulation_dataset_diff}
\end{table*}

\begin{table}[]
\centering\scriptsize
\scalebox{0.82}{
\begin{tabular}{@{}lclcl@{}}
\toprule
 & \multicolumn{2}{c}{\yoruba} & \multicolumn{2}{c}{Hausa} \\ \cmidrule(lr){2-3} \cmidrule(lr){4-5}
 & FT & \multicolumn{1}{c}{TP+FT} & FT & \multicolumn{1}{c}{TP+FT} \\ \midrule
Performance Difference (\%) & \multicolumn{1}{l}{1.93$\pm$1.71} & 1.00$\pm$0.70 & \multicolumn{1}{l}{1.08$\pm$0.79} & 1.92$\pm$1.64 \\ \bottomrule
\end{tabular}}
\caption{Average difference (in \%) and standard deviation (10 trials) between the test accuracy based on early-stopping on the clean validation and on the noisy validation set.}
\label{tab:realistic_result_diff}
\end{table}

\section{Comparing Early-stopping on Clean and Noisy Validation Sets}
\label{sec: appendix_compare_ec_en}
We compare the difference in model performance when using a noisy validation set rather than the clean one. Table \ref{tab:simulation_dataset_diff} presents the results on datasets with injected noise. For a noise level below 60\% uniform noise or 40\% single flip-noise, we see the difference is often less than 0.5\%, indicating that a noisy validation set can already serve as a good estimator for the generalization error. In an even higher noise level, the difference can be up to 2.14\%. As for the datasets obtained from weak supervision, the difference is higher in general. Table \ref{tab:realistic_result_diff} summarizes the difference on the \yoruba and Hausa.

\begin{table*}[]
\centering
\scalebox{0.60}{
\begin{tabular}{@{}lllllllllllll@{}}
\toprule
\multirow{3}{*}{} & \multicolumn{7}{c}{AG-News} & \multicolumn{4}{c}{IMDB} \\ \cmidrule(lr){2-8} \cmidrule(lr){9-12} 
 & \multicolumn{1}{c}{} & \multicolumn{3}{c}{uniform} & \multicolumn{3}{c}{single-flip} &  & \multicolumn{3}{c}{single-flip} \\ \cmidrule(lr){3-5} \cmidrule(lr){6-8} \cmidrule(lr){10-12} 
  & \multicolumn{1}{c}{clean} & \multicolumn{1}{c}{40\%} & \multicolumn{1}{c}{60\%} & \multicolumn{1}{c}{70\%} & \multicolumn{1}{c}{20\%} & \multicolumn{1}{c}{40\%} & \multicolumn{1}{c}{45\%} & \multicolumn{1}{c}{clean} & \multicolumn{1}{c}{20\%} & \multicolumn{1}{c}{40\%} & \multicolumn{1}{c}{45\%} \\ \midrule

NV & 94.07$\pm$0.13 & 84.48$\pm$0.78 & 61.61$\pm$3.18 & 43.78$\pm$5.07 & 90.46$\pm$0.37 & 76.06$\pm$0.33 & 64.74$\pm$0.94 & 94.03$\pm$0.13 & 86.34$\pm$0.77 & 65.05$\pm$0.90 & 58.97$\pm$1.26 \\
CT & - & 92.18$\pm$0.21 & 89.90$\pm$0.38 & 84.74$\pm$2.56 & 93.33$\pm$0.12 & 90.62$\pm$0.53 & 87.99$\pm$1.64 & - & 92.32$\pm$0.27 & 89.36$\pm$0.67 & 83.77$\pm$3.88 \\
NMat & - & 92.25$\pm$0.14 & 89.91$\pm$0.48 & 83.9$\pm$1.87 & 93.91$\pm$0.15 & 93.13$\pm$0.31 & 92.93 $\pm$0.51 & - & 92.07$\pm$0.21 & 87.13$\pm$0.44 & 78.82$\pm$1.37 \\
NMwR & 93.64$\pm$0.06 & 92.02$\pm$0.20 & 89.91$\pm$0.33 & 84.77$\pm$2.24 & 93.03$\pm$0.17 & 90.23$\pm$0.65 & 88.93$\pm$0.68 & 93.68$\pm$0.14 & 92.12$\pm$0.35 & 85.94$\pm$0.86 & 80.17$\pm$2.57 &\\
LS & 94.43$\pm$0.19 & 92.45$\pm$0.21 & 89.79$\pm$0.38 & 86.64$\pm$0.78 & 93.56$\pm$0.23 & 92.40$\pm$0.33 & 90.94$\pm$0.86 & 94.06 $\pm$0.09 & 92.13$\pm$0.43 & 87.22$\pm$1.39 & 80.61$\pm$2.48\\
WN & 94.40$\pm$0.13 & 92.40$\pm$0.25 & 89.53$\pm$0.75 & 85.49$\pm$0.76 & 93.80$\pm$0.08 & 92.33$\pm$0.35 & 88.94$\pm$0.92 & 93.98$\pm$0.15 & 92.13$\pm$0.21 & 85.88$\pm$2.78 & 80.12$\pm$4.09\\ \bottomrule
\end{tabular}}
\caption{Average test accuracy (\%) and standard deviation (5 trials) on AG-News and IMDB with uniform and single-flip noise. NV: without noise-handling and no validation set, i.e. train the model without noise-handling and until the training loss converges. CT: Co-teaching. NMat: Noise Matrix. NMwR: Noise Matrix with Regularization. LS: Label Smoothing. CT and NMat are equivalent to WN in the clean setting. Note that as IMDB is a binary-classification task, single-flip noise is equivalent to the uniform noise in this case.}
\label{tab:synthetic_result}
\end{table*}

\begin{table*}[]
\centering
\begin{tabular}{@{}lllll@{}}
\toprule
      & \multicolumn{2}{c}{\yoruba} & \multicolumn{2}{c}{Hausa} \\ \cmidrule(lr){2-3} \cmidrule(lr){4-5} 
      & \multicolumn{1}{c}{clean}        & \multicolumn{1}{c}{noisy}        & \multicolumn{1}{c}{clean}        & \multicolumn{1}{c}{noisy}        \\ \midrule
NV & 74.11$\pm$0.26   & 63.88$\pm$1.59 & 83.02$\pm$0.45 & 46.98$\pm$1.01                    \\
CT    & -           & 61.37$\pm$1.58 & - & 31.65$\pm$2.71                     \\
NMat  & -            & 65.96$\pm$0.81 & - & 46.58$\pm$0.88                      \\
NWwR  & 73.78$\pm$0.32           & 61.32$\pm$0.71 & 83.21$\pm$0.40 & 35.36$\pm$3.60                  \\
LS    & 74.22$\pm$0.37            & 65.44$\pm$1.67 & 83.44$\pm$0.35 & 46.44$\pm$0.78                 \\
WN    & 74.45$\pm$0.32            & 64.72$\pm$1.45 & 83.55$\pm$0.47
 & 46.97 $\pm$0.81                    \\ \bottomrule
\end{tabular}
\caption{Average test accuracy (\%) and standard deviation (10 trials) on \yoruba and Hausa with noise from weak supervision. NV: without noise-handling and no validation set, i.e. train the model without noise-handling and until the training loss converges. CT: Co-teaching. NMat: Noise Matrix. NMwR: Noise Matrix with Regularization. LS: Label Smoothing.}
\label{tab:realistic_result}
\end{table*}

\begin{figure}[t]
    \centering
    \includegraphics[width=0.4\textwidth]{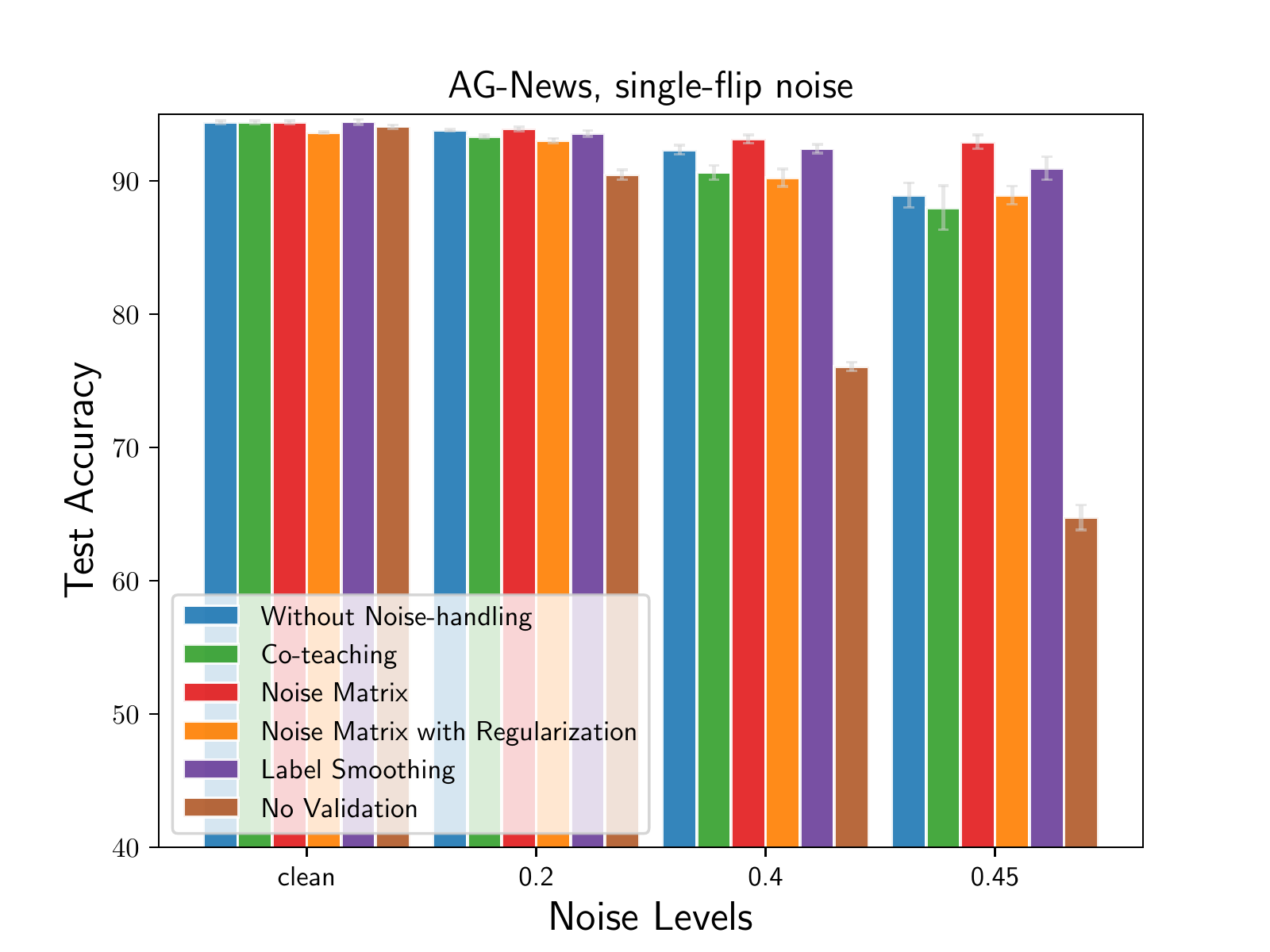}
    \caption{Test accuracy (\%) on AG-News dataset with single-flip noise. }
    \label{fig:appendix_agnews_sflip}
\end{figure}

\section{BERT Performance on Different Datasets and Noise Settings}
\label{sec:appendix_performance_table}
We evaluate the baselines under different noise settings and different datasets. The full result is shown in Table \ref{tab:synthetic_result} and Table \ref{tab:realistic_result}. A visualization of the result on AG-News with single-flip noise can be found in Figure \ref{fig:appendix_agnews_sflip} (other plots can be found in the main paper). BERT clearly shows its robustness against injected noise. Although noise-handling methods do help under a high noise level, the effect is limited (less than 4\%). Compared to injected noise, the noise from weak supervision is much more challenging for BERT, especially on the Hasua dataset. For both noise types, there is no single noise-handling method that outperforms the simple baseline method without noise-handling in all settings. 

\section{More ROC Curves}
\label{sec:appendix_more_roc_curves}
\begin{figure*}[t]
    \centering
    \subfigure[AG-News - 60\% uniform noise]{
        \includegraphics[width=0.65\columnwidth]{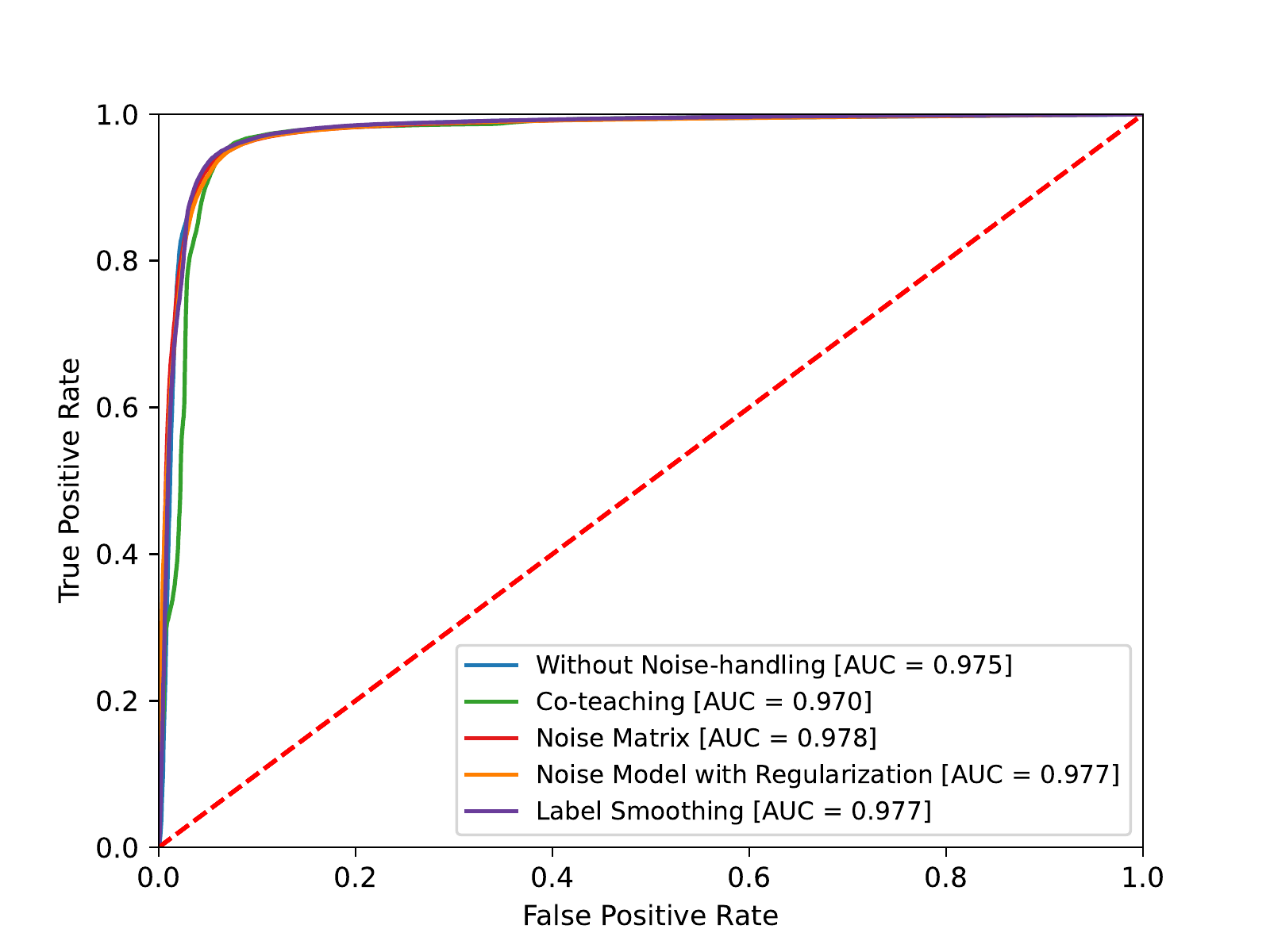}
    }
    \subfigure[AG-News - 40\% single-flip noise]{
        \includegraphics[width=0.65\columnwidth]{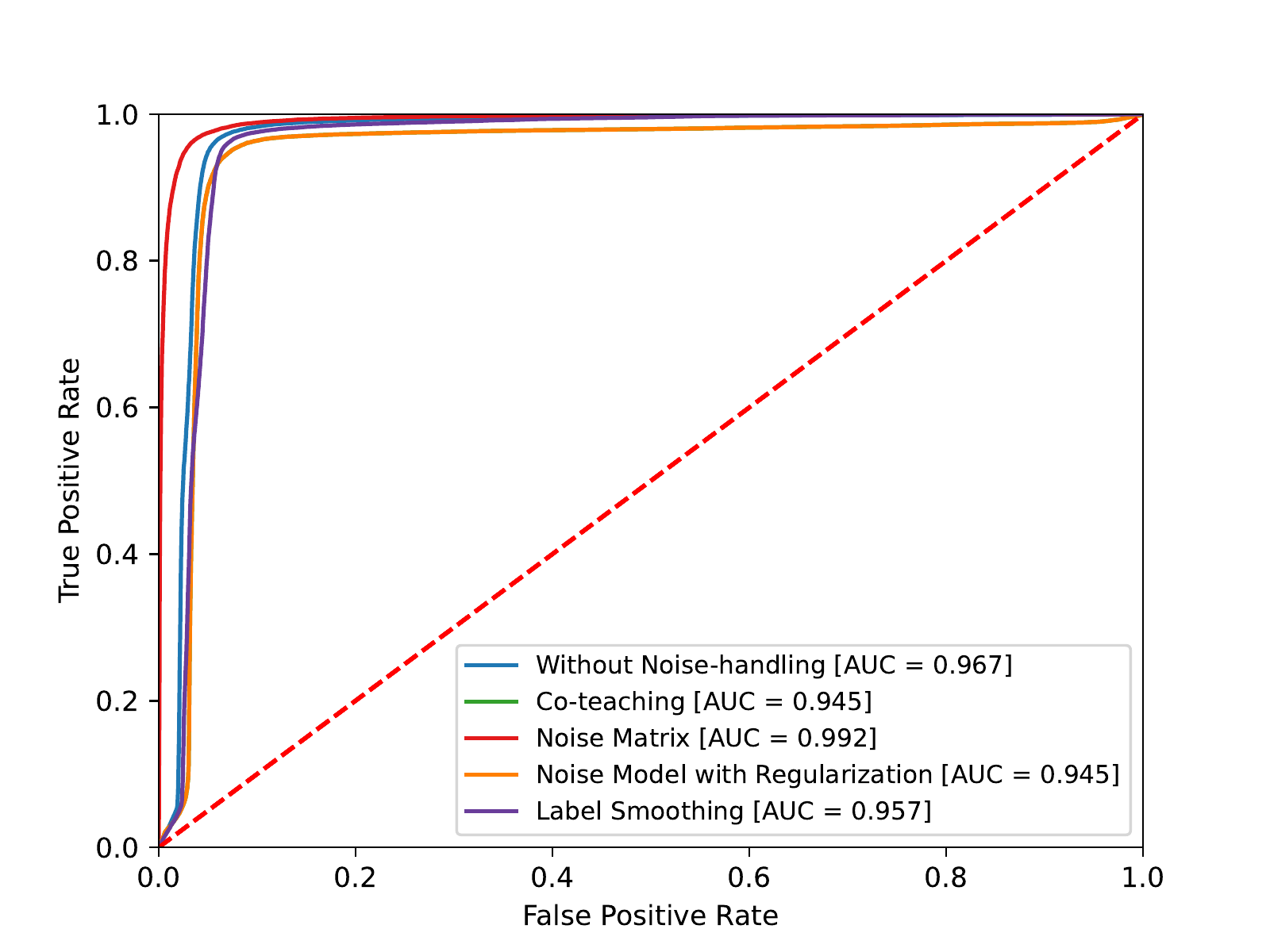}
    }
    \subfigure[AG-News - 45\% single-flip noise]{
        \includegraphics[width=0.65\columnwidth]{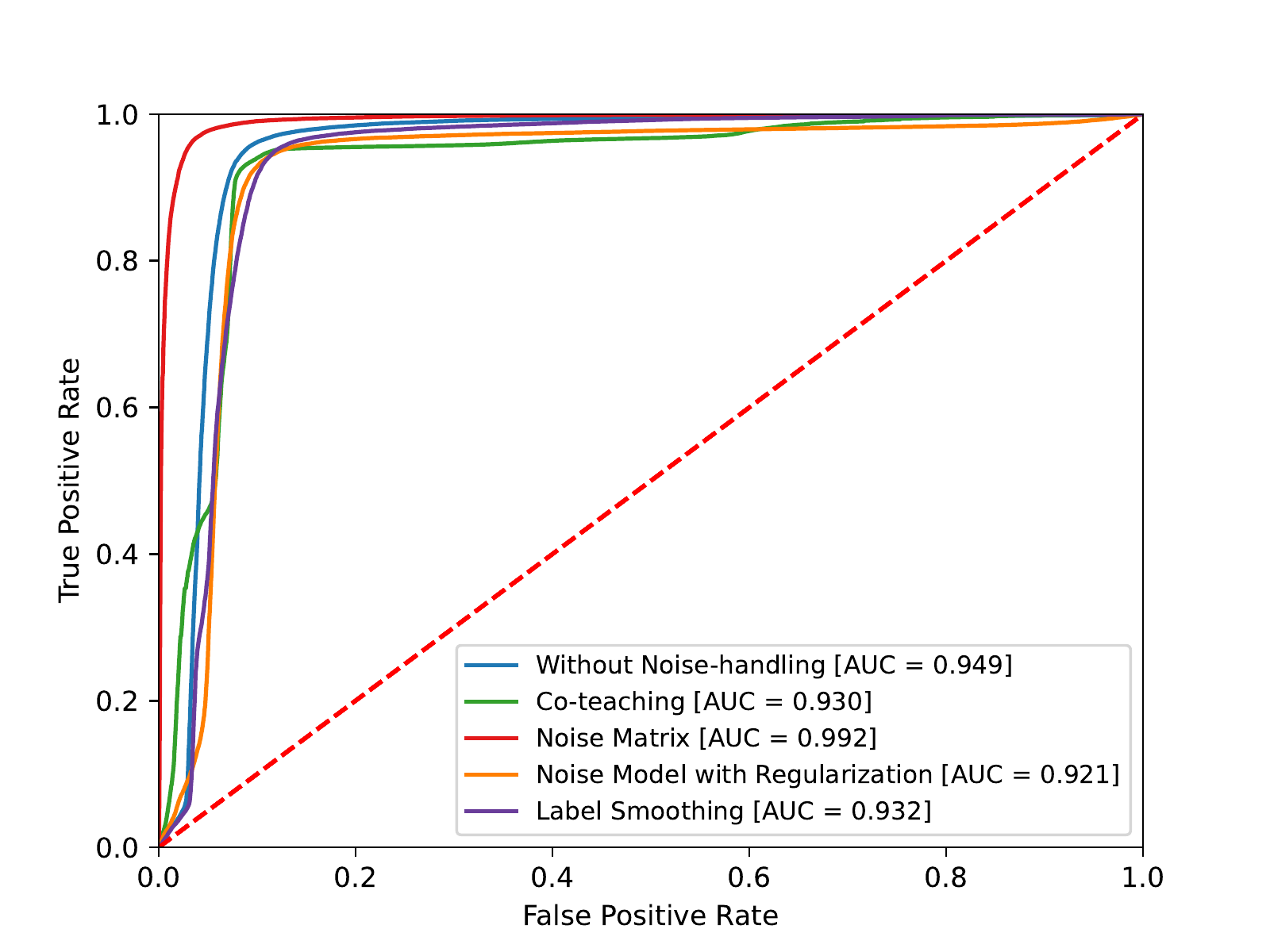}
    }
     \subfigure[IMDB - 40\% single-flip noise]{
        \includegraphics[width=0.65\columnwidth]{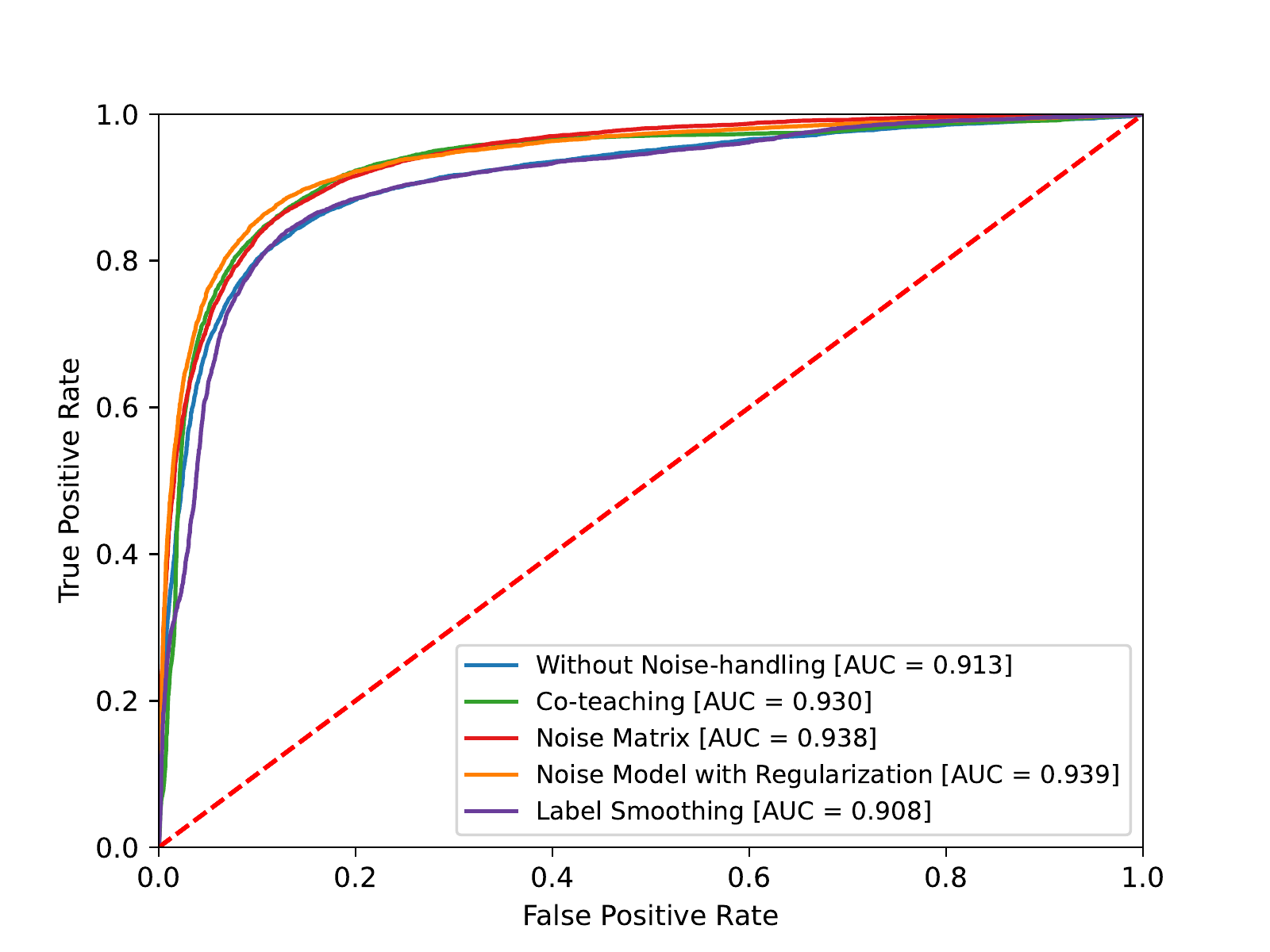}
    }
    \subfigure[IMDB - 45\% single-flip noise]{
    \includegraphics[width=0.65\columnwidth]{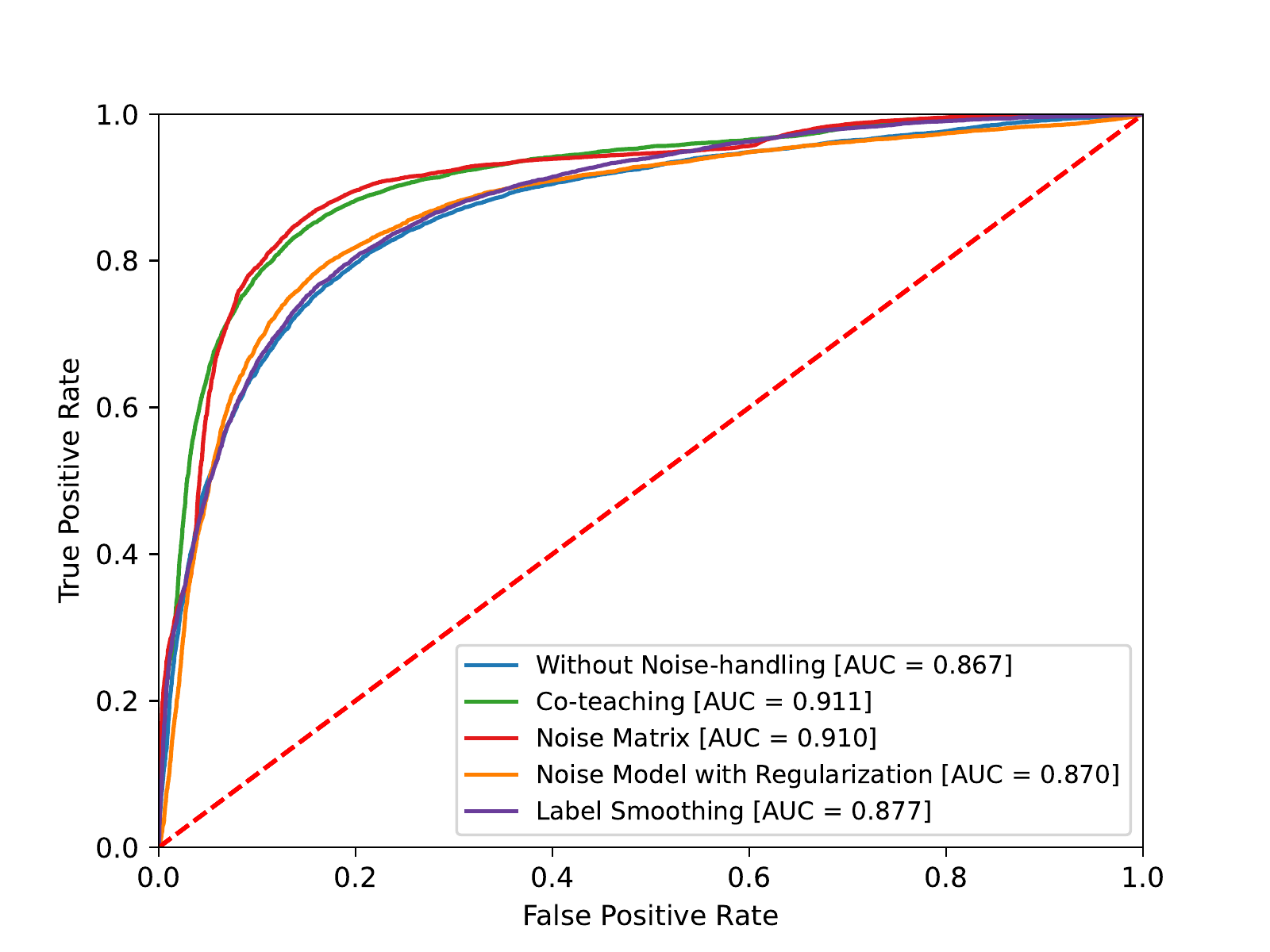}
    }
    \caption{The losses are record at the training step when early-stopping is triggered.  noise-handling methods do not make the losses of correct and incorrect labels more distinguishable. }
    \label{fig:more_roc_plots}
\end{figure*}

We present additional ROC curves under different settings with injected noise in Figure \ref{fig:more_roc_plots}. It is obvious that the AUC decreases when the noise levels increase. However, the absolute AUC score remains at a high level even under extremely high noise levels of injected noise.

\section{Implementation Details}
\label{sec: appendix_implementation_detail}

\paragraph{Datasets}
We experiment with the following four datasets: AG-News, IMDB, \yoruba and Hausa. 
\begin{enumerate}
  \item AG-News: originates from AG, which is a large collection of news articles. \citet{DBLP:conf/nips/ZhangZL15} constructed the AG-News dataset from the AG collection and it is used as a benchmark dataset for text classification.
  \item IMDB: consists of movie reviews with binary labels. It is a commonly used benchmark dataset used for text classification.
  \item \yoruba: The dataset was created from BBC \yoruba news titles along with the noisy dataset~\cite{DBLP:conf/emnlp/HedderichAZAMK20}. %
  \item Hausa: Similar to \yoruba, the Hausa dataset and the corresponding noisy dataset were created by \citet{DBLP:conf/emnlp/HedderichAZAMK20} from VOA Hausa news titles and distant supervision using keywords.  
\end{enumerate}

\paragraph{Models} We use the official BERT-base model \cite{DBLP:conf/naacl/DevlinCLT19} for text classification on AG-News and IMDB. It consists of an embedding layer, a 12-layer
encoder, and a pooling layer. It contains 110M parameters in total. We use the multilingual version of BERT-base for text classification on \yoruba and Hausa. It has the same architecture as the original BERT-base model. It also has 110M training parameters.

\paragraph{Fine-tuning on Text Classification Task} For the vanilla models (Wihtout Noise-handing and No Validation models in the paper), we pass the final layer of the [CLS] token representation ($\mathbb{R}^{768}$) to a feed forward layer for prediction. Noise Matrix and Noise Matrix with Regularization append a noise matrix $N \in \mathbb{R}^{k\times k}$ after the model's prediction. For Noise Matrix we initialize the matrix with the ground truth information. Following \cite{DBLP:conf/naacl/JindalPLN19}, when applying Noise Matrix with Regularization, we initialize the noise matrix using an identity matrix. The hyper-parameters for Noise Matrix with Regularization, Co-teaching and Label Smoothing are chosen so that the model performs the best on the noisy validation set. \\
In all settings, a batch size of 32 is used, and the learning rate is set to 2e-5. We train all models until the training loss converges. However, we report the score where the model performs the best on the validation set during training except for the No Validation baseline where we report the last-epoch performance.

\paragraph{Hardware and Average Runtime} We use Nvidia Tesla V100 and Nvidia GeForce GTX TITAN X to accelerate training. The average runtime for each method and dataset is summarized in Table \ref{tab:average_runtime}.

\begin{table}[]
\centering\scriptsize
\begin{tabular}{@{}lcccc@{}}
\toprule
 & \multicolumn{4}{c}{Average Runtime (Hours)} \\ \cmidrule(l){2-5} 
 & \multicolumn{1}{l}{AG-News} & \multicolumn{1}{l}{IMDB} & \multicolumn{1}{l}{\yoruba} & \multicolumn{1}{l}{Hausa} \\ \midrule
CT & 5 & 4.5* & 0.1* & 0.1* \\
NMat & 2.5 & 8 & 0.1* & 0.1* \\
NMwR & 3 & 8 & 0.1* & 0.1* \\
LS & 2.5 & 8 & 0.1* & 0.1* \\
WN & 2.5 & 8 & 0.1* & 0.1* \\\bottomrule
\end{tabular}
\caption{Average runtime (in hours) of each method. The Numbers with ``*'' indicates that the experiment was run on a Nvidia Tesla V100. Other experiments were run on a Nvidia GeForce GTX TITAN X.}
\label{tab:average_runtime}
\end{table}

\end{document}